\documentclass[conference]{IEEEtran}
\usepackage{times}

\usepackage[numbers]{natbib}
\usepackage{multicol}
\usepackage{tikz}    %
\usepackage{graphicx}
\usepackage[bookmarks=true]{hyperref}
\usepackage{multirow}
\usepackage{tcolorbox}
\usepackage{balance}

\usepackage{float} %

\newtcolorbox{examplebox}[2][]{
    colframe=black, colback=white, fontupper=\small\sffamily, fonttitle=\sffamily,
    boxrule=0.5pt, coltitle=white, title=#2, width=0.95\columnwidth, #1
}

\renewcommand{\paragraph}[1]{\vspace{1ex}\noindent\textbf{#1}}

\begin{document}

\title{Demonstrating the Octopi-1.5 Visual-Tactile-Language Model}

\author{\authorblockN{Samson Yu\authorrefmark{2}, Kelvin Lin\authorrefmark{2}, and Harold Soh\authorrefmark{2}\authorrefmark{3}
}
\authorblockA{\authorrefmark{2}Dept. of Computer Science, National University of Singapore}
\authorblockA{\authorrefmark{3}NUS Smart Systems Institute}
\authorblockA{\small Contact Authors: \texttt{samson.yu@u.nus.edu}, \texttt{harold@comp.nus.edu.sg}}
}

\maketitle

\begin{abstract}
Touch is recognized as a vital sense for humans and an equally important modality for robots, especially for dexterous manipulation, material identification, and scenarios involving visual occlusion. Building upon very recent work in touch foundation models, this demonstration will feature Octopi-1.5, our latest visual-tactile-language model. Compared to its predecessor, Octopi-1.5 introduces the ability to process tactile signals from multiple object parts and employs a simple retrieval-augmented generation (RAG) module to improve performance on tasks and potentially learn new objects on-the-fly. The system can be experienced live through a new handheld tactile-enabled interface, the TMI, equipped with GelSight and TAC-02 tactile sensors. This convenient and accessible setup allows users to interact with Octopi-1.5 without requiring a robot. During the demonstration, we will showcase Octopi-1.5 solving tactile inference tasks by leveraging tactile inputs and commonsense knowledge. For example, in a Guessing Game, Octopi-1.5 will identify objects being grasped and respond to follow-up queries about how to handle it (e.g., recommending careful handling for soft fruits). We also plan to demonstrate Octopi-1.5's RAG capabilities by teaching it new items. With live interactions, this demonstration aims to highlight both the progress and limitations of VTLMs such as Octopi-1.5  and to foster further interest in this exciting field. Code for Octopi-1.5 and design files for the TMI gripper are available at \url{https://github.com/clear-nus/octopi-1.5}.
\end{abstract}

\section{Introduction}

Touch has long been recognized as crucial for robots, particularly in contact-rich tasks or situations with visual occlusion. It enables robots to discern latent object properties, such as determining the softness or deformability of objects. In recent years, robot touch has made significant strides, with notable advances in sensing technologies --- for example, in  visual tactile sensors like GelSight~\cite{yuan2017gelsight} and Digit~\cite{lambeta2020digit} --- and in the interpretation and application of tactile signals for manipulation. Notably, recent work has integrated tactile perception with Vision-Language Models (VLMs), enabling robots to perform various tactile-related tasks using natural language prompts~\cite{yu2024octopi,tvl2024,yang2024binding}.

\begin{figure}
\centering
\includegraphics[width=0.5\textwidth]{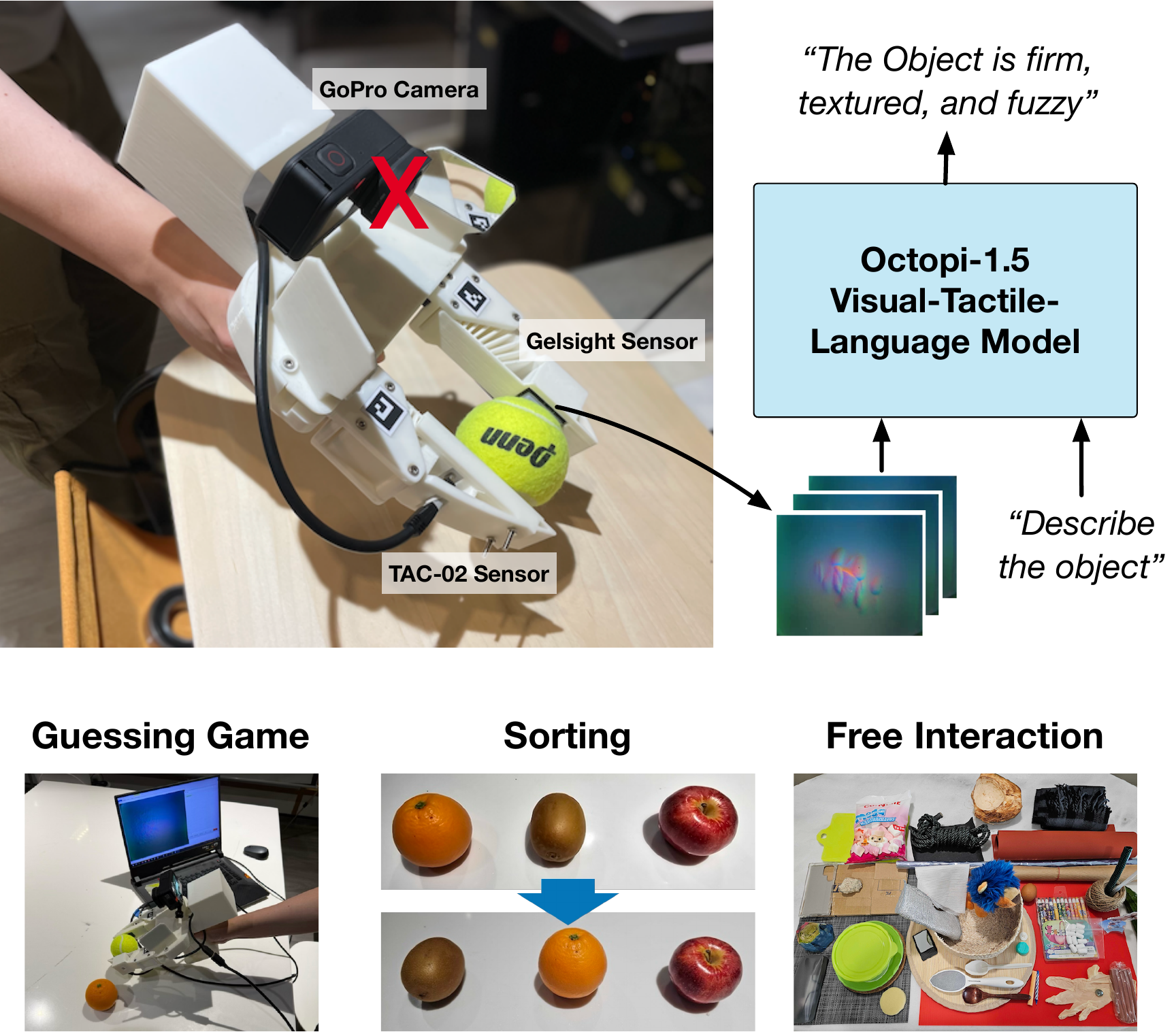}
\caption{Octopi-1.5 Demonstrations using a Tactile Manipulation Interface (TMI) gripper. We plan to demonstrate how Octopi-1.5 can be used to describe and make use of tactile sensations in a range of scenarios, e.g., without vision as shown in the top image. We also a (i) \textbf{Guessing Game} where Octopi-1.5 is tasked to guess which item is being grasped given only tactile inputs (without vision), (ii) a \textbf{Sorting} task where Octopi-1.5 has to sort items according to hardness or fruits according to ripeness, and (iii) a \textbf{Free Interaction} setting where users can use visual, tactile, and language modalities together with Octopi-1.5. in a free manner on a variety of objects to explore the strengths and limitations of the system.}
\label{fig:demos}
\end{figure}

Building upon this line of research, we have developed an improved visual-tactile language model (VTLM), which we call Octopi-1.5. Compared to existing VTLMs, Octopi-1.5 features three key enhancements: 
\begin{itemize}
    \item a new tactile encoder trained on a combination of existing Gelsight datasets with an expanded PhysiCLeAR dataset~\cite{yu2024octopi},
    \item the use of the Qwen2-VL 7B base VLM~\cite{Qwen2VL}, offering enhanced interaction and commonsense knowledge capabilities, the visual modality, and 
    \item experimental retrieval-augmented generation (RAG)~\cite{gao2023retrieval}, which not only boosts performance but potentially allows the model to learn new tactile-object pairings by adding them to its database.
\end{itemize}

At R:SS 2025, we propose to demonstrate Octopi-1.5’s capabilities in real-time using a 
Tactile Manipulation Interface (TMI) (Fig. \ref{fig:demos}), which is a Universal Manipulation Interface (UMI)~\cite{chi2024universal} modified with tactile sensors. The demo setup will allow users to interact with objects using the TMI and query Octopi-1.5 through natural language and physical interaction. We designed the setup to be portable and not require physical robot, which eases transportation to R:SS 2025.  We envision several interactive scenarios, including a Guessing Game --- in which Octopi-1.5 must identify the object being touched from a collection of items --- to free interaction where users can interact freely with Octopi and the TMI.

\begin{figure*}
\centering
\includegraphics[width=0.95\textwidth]{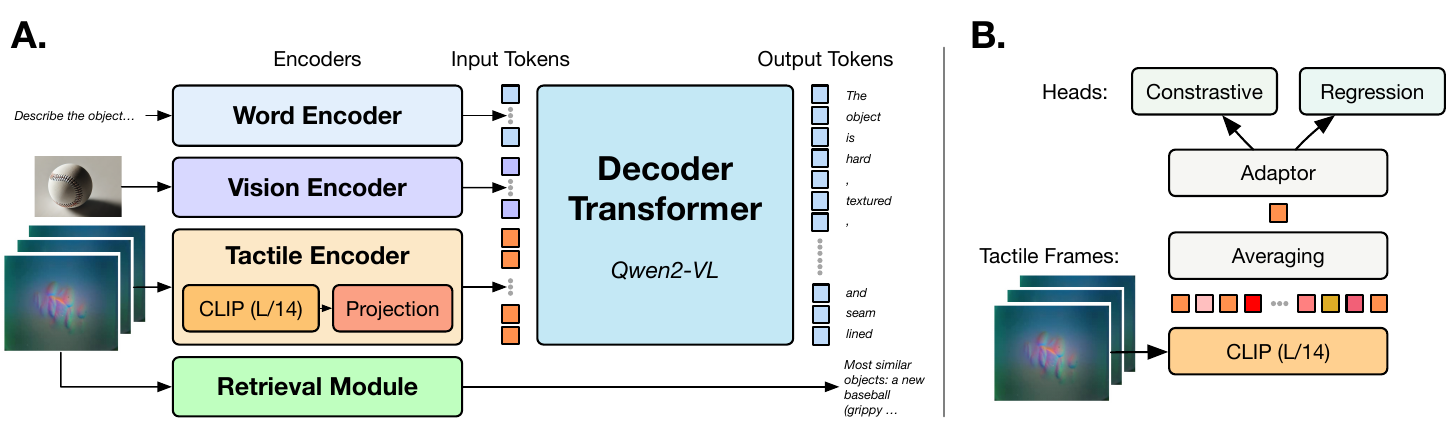}
\caption{(\textbf{A}) Octopi-1.5 model architecture. Octopi-1.5 is a fine-tuned Qwen2-VL 7B multimodal VLM with a tactile encoder and retrieval module. (\textbf{B}) The CLIP module of the tactile encoder is pre-trained before finetuning using contrastive and regression losses on PhysiCLeAR, Hardness, and ObjectFolder datasets. Please see main text for details.}
\label{fig:octopiarch}
\end{figure*}

The remainder of this paper details Octopi-1.5 and the demonstration setup. We begin with a brief review of related work in tactile-language models (Sec. \ref{sec:litreview}), followed by a description of Octopi-1.5’s model architecture, training process, and the TMI (\ref{sec:system}). Sec. \ref{sec:demos} outlines the planned demonstrations, supported by preliminary experimental results that validate their feasibility. Finally, we conclude with a summary and a discussion of current limitations (Sec. \ref{sec:conclusions}).

Our demonstrations aim to showcase advancements in tactile-enabled VLMs and their potential applications in robotics. While Octopi-1.5 and similar models are still under development, current limitations, such as challenges in generalization, remain areas of active research. We hope the demonstration will inspire discussion and further exploration in this field. All code for Octopi-1.5 and the TMI gripper design is available at \url{https://github.com/clear-nus/octopi-1.5}.

\section{Background and Related Works}
\label{sec:litreview}

Octopi-1.5 builds upon a wide base of prior work in tactile perception for robotics and multi-modal language models. In the following, we give a brief overview of these areas, focusing on advances in visual-tactile-language models. We refer readers interested in general tactile sensing and perception for robotics to survey articles~\cite{LUO201754,roberts2021soft}.

\paragraph{Tactile Sensing.} Although not as well-developed as vision, there has been significant progress on the development of tactile sensors over the years~\cite{li2024comprehensive}. Tactile sensors come in a variety of types depending on their sensing modality, e.g., piezoresistive, capacitive, and optical. In this work, we primarily use an optical sensor called the Gelsight Mini~\cite{yuan2017gelsight}\footnote{Gelsight Mini Product Sheet: \url{https://www.gelsight.com/wp-content/uploads/productsheet/Mini/GS_Mini_Product_Sheet_10.07.24.pdf}}. Optical tactile sensors provide high-resolution tactile images that capture surface deformations by converting them into visual data~\cite{li2024comprehensive}. This makes them  effective at detecting fine surface features and texture, and their image outputs are easily processed by modern machine learning methods. However, compared to alternative sensor types, optical tactile sensors tend to be larger in size and have lower sampling rates. 

Our TMI gripper also incorporates a piezoresistive sensor called the TAC-02\footnote{\url{https://www.tacniq.ai/tac-02-robotic-finger-dev-kit}}. Unlike optical sensors, piezoresistive sensors operate by detecting changes in resistance caused by the deformation of elastic materials under applied forces~\cite{li2024comprehensive}. The TAC-02 has 64 taxels, which is lower in resolution compared to the Gelsight Mini, but is more compact, has a higher sampling rate (1kHz), and can directly capture static and dynamic pressure.

\paragraph{Visual-Tactile-Language Models.} Octopi is closely related to very recent work on multimodal large-language models (MLLMs)~\cite{mllmsurvey} that process data from real-world tactile sensors~\cite{yu2024octopi,tvl2024,yang2024binding} --- we refer to these models as Visual-Tactile-Language models (VTLMs). These prior works share similarities in that they involve the collection of tactile, visual, and language modalities into datasets. For example, the TVL Dataset~\cite{tvl2024} includes over 44,000 paired tactile-visual samples annotated with natural language, while other datasets, such as ObjectFolder~\cite{gao2021ObjectFolder} and PhysiCLeAR~\cite{yu2024octopi}, combine high-resolution tactile data with semantic labels for material properties like hardness and roughness. These datasets are used to train tactile encoders, often based on Visual Transformers (ViTs)~\cite{dosovitskiy2021an}, employing contrastive and regression-based loss functions to align tactile, visual, and language representations. Octopi-1.5 shares these foundational approaches but (i) is trained on a larger dataset compared to its predecessor and (ii) incorporates a retrieval-augmented generation (RAG) module, enabling it to retrieve and utilize similar objects from its database to enhance predictions and allows for on-the-fly learning of new object-tactile pairings.

\begin{table}
\centering
\caption{Octopi-1.5 Training Dataset Statistics}
\begin{tabular}{l|c|c|c}
\hline\hline 
\textbf{Dataset}       & \textbf{Num.} & \textbf{Samples} & \textbf{Num. Tactile} \\ 
                       & \textbf{Objects}     & \textbf{per Object} & \textbf{Videos} \\ \hline\hline 
PhysiCLeAR-Plain       & 100                   & 5--45              & 2689                \\ 
PhysiCLeAR-Dotted      & 68                    & 11--32             & 1939                \\ \hline 
Hardness~\cite{yuan2016estimating}               & 210                   & 1--133             & 1860                \\ 
ObjectFolder-Real~\cite{gao2021ObjectFolder}      & 100                   & 30--67             & 3550                \\ \hline\hline
\end{tabular}
\label{tbl:datasets}
\vspace{-2em}
\end{table}

\begin{table*}[ht]
\centering
\caption{Annotator Scoring Guidance for Hardness and Roughness.}
\begin{tabular}{l|c|p{6cm}|l}
\hline\hline
\textbf{Property} & \textbf{Score} & \textbf{Description} & \textbf{Example Object} \\ \hline\hline
\multirow{3}{*}{Hardness} & 0  & Easily compressible when pressed with little force & Cotton wool \\ 
                          & 5  & Compressible upon pressing with moderate force & Foam mattress \\ 
                          & 10 & Incompressible with high human pressing force & Rock \\ \hline
\multirow{3}{*}{Roughness} & 0  & Minimal feeling of friction upon finger sliding & Ice \\ 
                           & 5  & Slight feeling of friction upon finger sliding & Jeans \\ 
                           & 10 & Significant feeling of friction upon finger sliding & Toothbrush bristles \\ \hline\hline
\end{tabular}
\label{tbl:hardness_roughness}
\vspace{-1em}
\end{table*}

\begin{figure}
\centering
\begin{examplebox}[label=taskexample]{Training Task Example: Describe and Rank}
\textbf{USER}\\
\texttt{Describe the objects in the following tactile videos and rank them in decreasing hardness and roughness.}\\[1ex]

\texttt{Object 1}\\
\texttt{Part 1.1: [tactile frames]}\\
\texttt{Part 1.2: [tactile frames]}\\[2ex]

\textbf{OCTOPI-1.5}\\
\texttt{Object 1}\\
\texttt{Part 1.1: [descriptions]}\\
\texttt{Part 1.2: [descriptions]}\\[1ex]

\texttt{Object parts ranked in decreasing hardness: 1.1, 1.2}\\
\texttt{Object parts ranked in decreasing roughness: 1.2, 1.1}
\end{examplebox}
\caption{Training Task Example. Octopi-1.5 is trained to describe and rank different parts of an object (in addition to different objects).}
\label{taskexample}
\vspace{-1em}
\end{figure}

\begin{figure}
\centering
\begin{examplebox}[label=RAGexample]{RAG Description Example}
\textbf{USER}\\
\texttt{Describe the object in the following tactile video(s).}\\[1ex]

\textbf{OCTOPI-1.5}\\
\texttt{Object 1: [Tactile Frames]}\\[2ex]

\texttt{Object 1: textured, grippy, seam-lined, hard.}\\
\textcolor{blue}{\texttt{Most similar objects (in order of decreasing similarity): a new baseball's seams (grippy, hard, seam-lined, textured); ...}}
\end{examplebox}
\caption{The outcome of a tactile description by Octopi-1.5 augmented with RAG. The additional information added by the RAG module is shown in \textcolor{blue}{blue}.}
\label{fig:RAGexample}
\end{figure}

\section{System Description: Octopi-1.5 and TMI}
\label{sec:system}
In this section, we discuss the main components of our demonstration: Octopi-1.5 and the TMI. 

\subsection{Octopi-1.5 VTLM}

Octopi-1.5 is a visual-tactile-language model (VTLM). Below, we present the model’s architecture, training methodology, and the integration of a simple RAG module.

\paragraph{Octopi-1.5 Model Structure.}
Fig. \ref{fig:octopiarch}.A provides a high-level overview of the model architecture, including the GelSight mini tactile encoder. Octopi-1.5 is based on the QWEN2-VL 7B open-source vision-language model (VLM)~\cite{Qwen2VL}, whereas the previous Octopi version utilized Llama~\cite{zheng2023judging}. Encoders are used to transform raw inputs into tokens, which are subsequently processed by the VLM transformer.

A key improvement in Octopi-1.5 is in the tactile encoder, which translates tactile frames from a GelSight mini sensor into tokens for the VLM. Specifically, the tactile encoder is a fine-tuned CLIP~\cite{radford2021learning} module augmented with a projection layer. To optimize computational efficiency, we process only ``salient'' frames, which are selected using a heuristic that identifies the top 10 frames with the largest differences compared to their preceding frames.

\paragraph{Octopi-1.5 Training.}
Octopi-1.5 was trained on an expanded PhysiCLeAR dataset (including both marker and markerless GelSight pads), as well as the hardness~\cite{yuan2016estimating} and ObjectFolder datasets~\cite{gao2021ObjectFolder}. Table \ref{tbl:datasets} provides a summary of the dataset statistics. Model training was conducted in {two stages}: first, we trained the CLIP module within the tactile encoder, followed by end-to-end fine-tuning of the VTLM. This staged approach resulted in more stable training outcomes.

The CLIP module was trained using both regression and contrastive losses, as illustrated in Fig. \ref{fig:octopiarch}.B. In the current implementation, the network is trained to predict human-annotated hardness and roughness scores for objects. Annotators followed the guidance shown in Table \ref{tbl:hardness_roughness} for labeling. For the contrastive loss, the network distinguishes tactile inputs from the same object/part (positive class) versus those from other objects/parts (negative class). To ensure consistency, both positive and negative classes were sampled using the same type of GelSight silicon pad (marker or markerless). We trained the CLIP module for 30 epochs, selecting the best encoder based on validation loss using a set of six unseen objects. 

In the second phase, the CLIP module is frozen, and the projection layers are trained alongside the decoder. The decoder training employs Visual Prompt Tuning (VPT)~\cite{jia2022visual} on description and ranking tasks as illustrated in Fig.~\ref{taskexample}. Unlike its predecessor, Octopi-1.5 is explicitly designed to handle objects comprising multiple parts, such as a hairbrush with distinct handle and bristle components.

\paragraph{RAG-Modified Descriptions.} We are currently experimenting with the decomposition of a VTLM into two components: one dedicated to processing/inference and another serving as a ``memory'' of previously seen items. As a step towards this goal, Octopi-1.5 incorporates a simple RAG scheme to enhance its tactile descriptions by augmenting them with textual information from similar objects, thereby aiding downstream tasks. Specifically, the tactile descriptions of objects are supplemented with their labels and the tactile descriptions of similar items, creating a more informative representation (example in Fig.~\ref{fig:RAGexample}).

To generate these augmented descriptions, we follow a straightforward process:
\begin{itemize}
    \item For a new set of salient tactile images, compute their average embedding using the tactile encoder.
    \item Perform a cosine-similarity search over an existing dataset of tactile embeddings to identify the top-5 objects with the most similar averaged embeddings.
    \item Aggregate the unique objects and rank them by the number of retrieved samples to form a prioritized list of matches.
\end{itemize}
While straightforward, this RAG scheme improved Octopi's performance in our proposed demonstration tasks (see Section \ref{sec:demos}). Future work could explore more advanced RAG setups, such as retrieving and incorporating tactile embeddings directly; implementing such improvements will likely require retraining the VTLM to optimally utilize the retrieved information.

\begin{figure}
    \centering
    \includegraphics[width=0.45\textwidth]{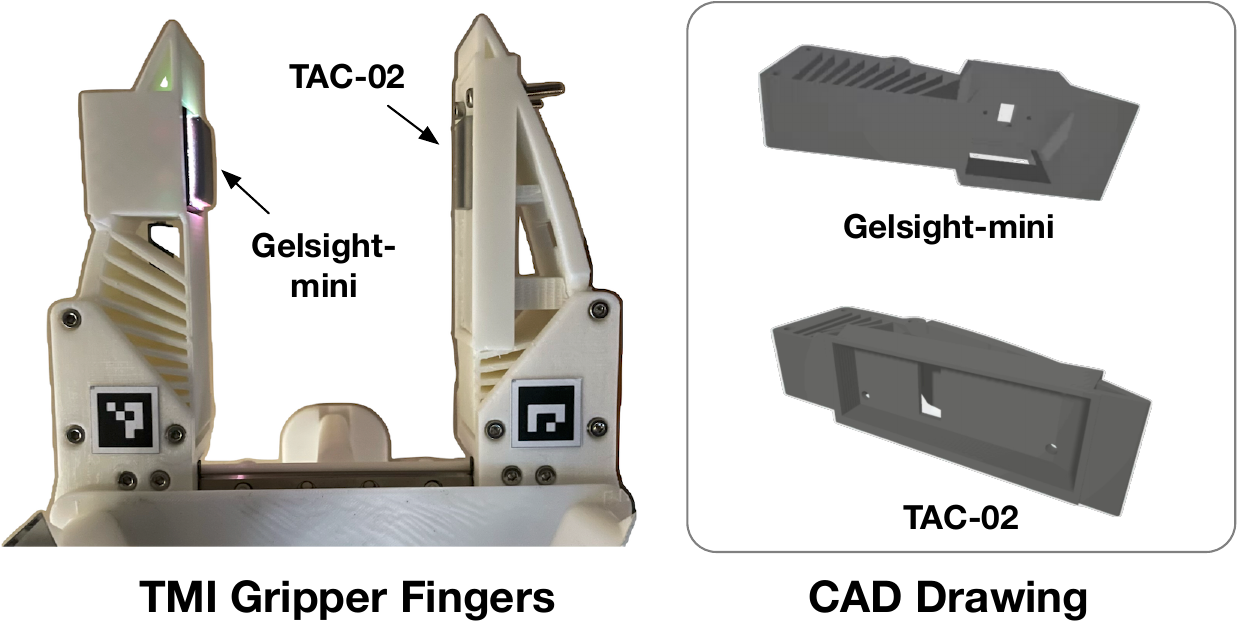}
    \caption{TMI fingers and CAD drawings with compartments for inserting the Gelsight-mini or the TAC-02 sensors.}
    \label{fig:TMIFingers}
\end{figure}

\subsection{Tactile Manipulation Interface} 
To facilitate the demonstration, we propose using a modified Universal Manipulation Interface (UMI)~\cite{chi2024universal}. Our Tactile Manipulation Interface (TMI) features fingers equipped with tactile sensors (Fig.~\ref{fig:TMIFingers}). Specifically, one finger is mounted with a GelSight Mini sensor, while the other houses a TAC-02 piezoresistive tactile sensor. The GelSight sensor provides high-resolution tactile imaging, whereas the TAC-02 sensor~\footnote{\url{https://www.tacniq.ai/tac-02-robotic-finger-dev-kit}} directly captures pressure readings.

Our demonstration will primarily utilize the GelSight Mini sensor, as it has significantly more data available for model training. Nevertheless, the TMI provides a proof-of-concept integration of multiple sensors, which we believe will spur discussion on integrating heterogeneous tactile modalities. We also hope to show participants how manipulation data can be collected using the TMI. We plan to engage in broader discussions surrounding TMI's benefits --- such as ease of use and versatility --- and limitations including sensor wear and tear, task constraints, difficulties in annotation, and the variability across different tactile sensors.

\begin{figure}
    \centering
    \includegraphics[width=0.45\textwidth]{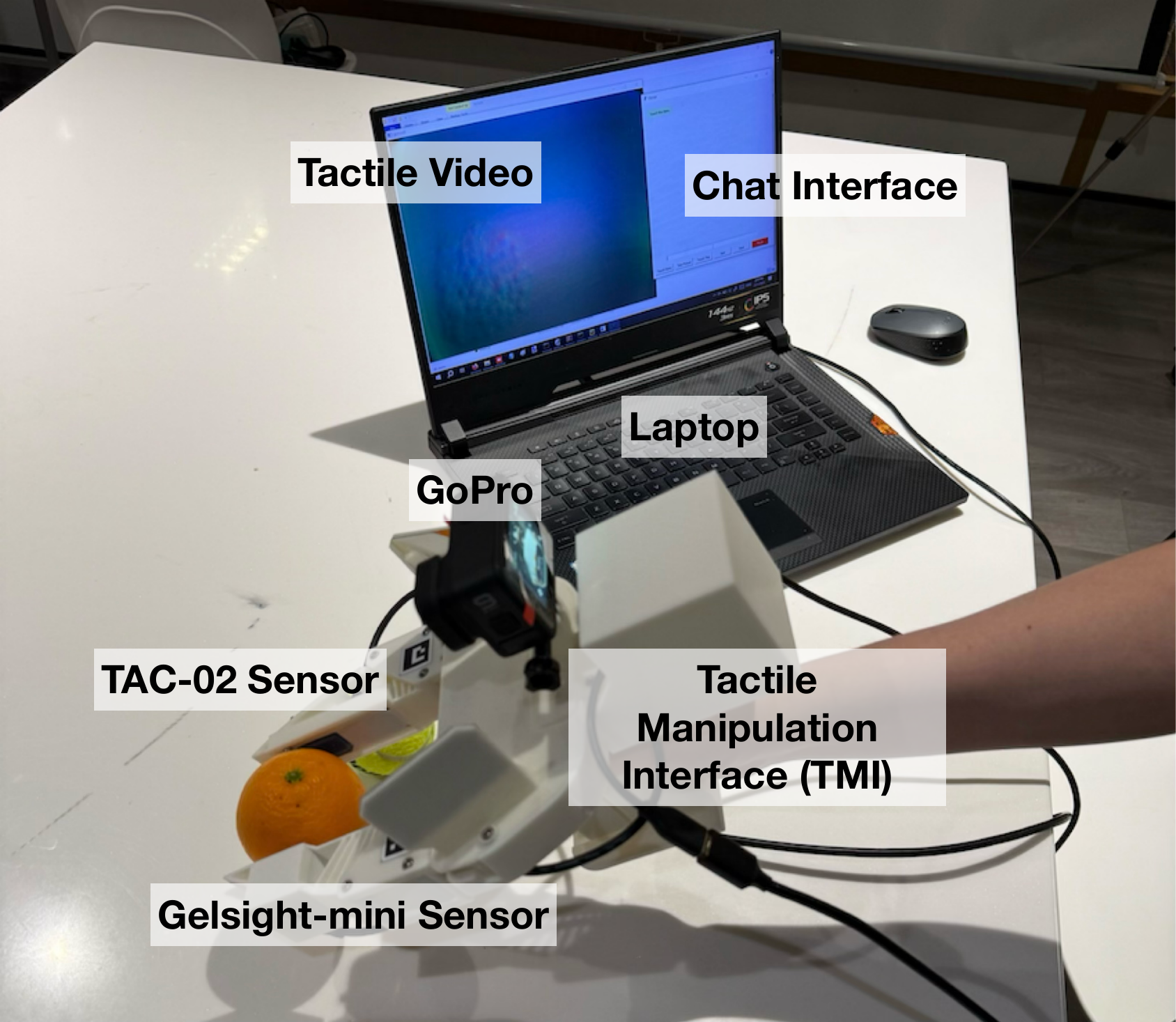}
    \caption{Our demonstration system is highly portable, consisting primarily of the TMI and a laptop. The laptop is equipped with the necessary software to interface with Octopi-1.5, which runs on a remote workstation or server. A stable internet connection is required, with the option to use a mobile hotspot if needed. The setup can be fully assembled in under 15 minutes. We will provide a variety of items for grasping and participants can use their own objects.}
    \label{fig:demosetup}
\end{figure}

\begin{figure}
\centering
{\includegraphics[width=0.20\textwidth]{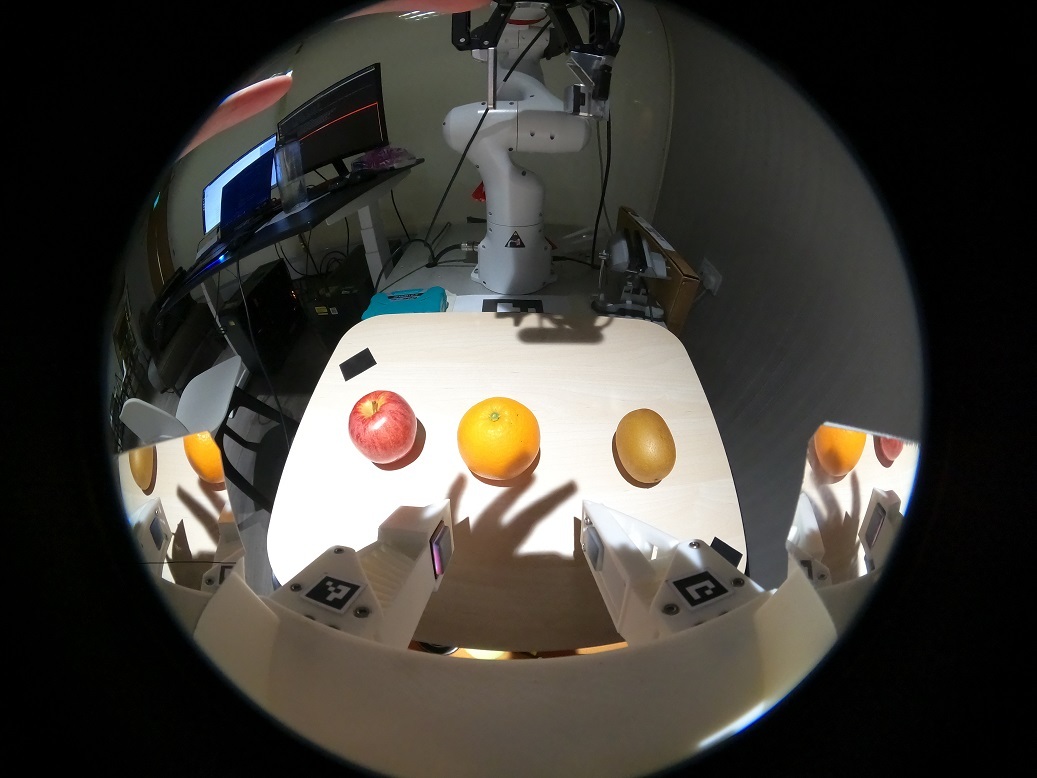}
}
\begin{examplebox}[]{Guessing Game Example (through Simplified Interface)}
{\small
\textbf{USER}\\
\texttt{Take a picture of the scene.}\\
\texttt{[System takes a picture using the GoPro; see image above].}\\
\texttt{What fruits do you see in the image?}\\[1ex]

\textbf{OCTOPI-1.5}\\
\texttt{Object 1: round, red, apple.}\\
\texttt{Object 2: round, orange, orange.}\\
\texttt{Object 3: oval, yellow, kiwi.}\\[1ex]

\textbf{USER}\\
\texttt{Touch an item on the table.}\\[1ex]

\textbf{OCTOPI-1.5}\\
\texttt{Please use the gripper to grab the item.}\\
\texttt{[User grasps apple using the TMI (no visual images taken)].}\\
\texttt{Finished collecting tactile data.}\\

\textbf{USER}\\
\texttt{Describe what you just touched.}\\[1ex]

\textbf{OCTOPI-1.5}\\
\texttt{Object 4: hard, firm, glossy, smooth.}\\[1ex]

\textbf{USER}\\
\texttt{Which fruit is it likely to be?}\\[1ex]

\textbf{OCTOPI-1.5}\\
\texttt{To determine which option the object is likely to be, we need to consider the surface texture descriptions that help to distinguish between the given options. Let's analyze each option:\\
A) Apple: Apples typically have a smooth, slightly waxy texture on the skin.\\
B) Orange: Oranges have a rough, bumpy texture on the skin.\\
C) Kiwi: Kiwis have a fuzzy, textured skin.\\
Now, let's give a succinct case for each option ...\\
Answer: (A) Apple is the most likely option because it has a smooth, slightly waxy texture on the skin, which is characteristic of apples.}}
\end{examplebox}
\caption{User Interaction Example for the Guessing Game.}
\label{fig:GuessExample}
\end{figure}

\section{Planned Demonstrations}
\label{sec:demos}
We propose a series of demonstrations to showcase Octopi-1.5's capabilities in aiding tactile-related tasks. These demonstrations will utilize the TMI, connected to a laptop with an internet connection to a remote workstation/server at NUS running Octopi-1.5 (Fig. \ref{fig:demosetup}). If available, we would prefer to use a large-screen TV or projector to display presentation slides and Octopi-1.5's feedback.

Our planned format is straightforward: we will begin with a brief overview of Octopi-1.5 and the TMI, followed by an introduction to the demonstration system. Currently, text interaction is facilitated through a simple chat interface, though we are exploring the feasibility of integrating speech-to-text using Whisper~\cite{radford2023robust}. However, we found Whisper can be unreliable in noisy environments with multiple speaking participants, so it may not be included in the final setup. 

We will then demo a tactile-based description task (Fig.~\ref{fig:demos}), followed by two structured demonstrations: the Guessing Game and Sorting Task. Finally, users will have the opportunity to interact more freely with the system to explore its capabilities and limitations.

\begin{table}
\centering
\caption{Average accuracies (\%) on the Guessing Game. }
\begin{tabular}{l  c c c c c}
\hline\hline
\textbf{Method} & \textbf{RAG} & \textbf{Balls} & \textbf{Fruits} & \textbf{Unseen} & \textbf{Unseen} \\
  &  &  &  &  & \textbf{(teaching)}
\\ 
\hline\hline
Encoder-1.5      & --   & 80.00  & 100.00 & N/A    & 89.02 \\
\hline 
Octopi-1 (7B)       & No   & 44.00  & 42.31  & 43.90  & N/A   \\
+-Octopi-1 (13B)     & No   & 48.00  & 34.62  & 53.66  & N/A   \\
Octopi-1.5 (8B)    & No   & 56.00  & 57.69  & 41.46  & N/A   \\
Octopi-1.5 (8B)     & Yes  & 96.00  & 100.00 & 73.17  & 95.12 \\
\hline\hline
\end{tabular}
\label{tbl:guess_game_perf}
\end{table}

\subsection{Guessing Game}
The guessing game is designed to demonstrate how tactile information can be combined with commonsense knowledge to make inferences:
\begin{itemize}
    \item We will provide octopi with a set of objects (described via language or visual input)
    \item Users will select an object and provide tactile inputs by grasping the object with the TMI (no visual input provided during grasping)
    \item Octopi-1.5 will infer which object corresponds to the tactile inputs.
    \item Users can interact with the system by prompting Octopi-1.5 to guess again if the initial guess was incorrect or asking follow-up questions (e.g., how hard the object should be grasped).
\end{itemize}
To simplify the interaction, the demo will make use of a chat interface that ``hides'' the more complex prompts used to elicit proper behavior from Octopi-1.5; see Fig. \ref{fig:GuessExample}) for an example of the interaction where the user grasps an apple. After the interaction, we will also show the exact prompts that are used ``behind-the-scenes''.

Preliminary tests support the feasibility of this demonstration. We experimented with three categories of objects: balls (baseball, plush ball, tennis ball) and fruits (apple, partially ripe kiwi, orange) and completely unseen objects (hairbrush bristles, hairbrush handle, microfiber cloth). Accuracy scores are summarized in Table~\ref{tbl:guess_game_perf}, with averages computed across multiple samples from the same object (25 samples for balls, 26 samples for fruits, and 41 unseen samples).

As a non-VLM baseline, the Encoder-1.5 model uses cosine similarity of tactile embeddings in the training set. The encoder-only baseline achieves good results on familiar objects but offers no out-of-the-box handling of unseen items without database updates. 

Octopi-1.5 shows higher overall accuracy compared to the older Octopi-1. Simple {RAG} retrieval improves performance and ``teaching'' further boosts results by adding new item samples to the dataset on the fly.  With RAG enabled, accuracy scores for Octopi-1.5 are high (96-100\%) for the seen objects (Balls, Fruits) with good performance (73\%) for unseen objects. Once teaching is enabled, performance rises to (95\%). 
Along with these items, we hope the demonstration will also show Octopi-1.5's ability to recognize multiple parts of the same object.

\begin{figure}
\centering
\begin{examplebox}[label=SortExample]{Sorting Example with RAG (through Simplified Interface)}
{\small
\textbf{USER}\\
\texttt{Touch this item.}\\[1ex]

\textbf{OCTOPI-1.5}\\
\texttt{Please use the gripper to grab the item.}\\
\texttt{[User grasps apple using the TMI (no visual images taken)].}\\
\texttt{Finished collecting tactile data.}\\[1ex]

[... Repeated 3 times...] \\[1ex]

\textbf{USER}\\
\texttt{Describe and rank the objects by their hardness.}\\[1ex]

\textbf{OCTOPI-1.5}\\
\texttt{Object 1: smooth, hard, slightly waxy.\\
\textcolor{blue}{Most similar objects (in order of decreasing similarity): an unpeeled, ripe apple (firm, glossy, slightly waxy, smooth); the back of a TV remote (hard, slightly grainy, smooth);}\\
Object 2: textured, slightly soft, rounded.\\
\textcolor{blue}{Most similar objects (in order of decreasing similarity): an unpeeled, ripe orange (rounded, slightly soft, textured);}\\
Object 3: slightly rough, fuzzy, squishy, soft.\\
\textcolor{blue}{Most similar objects (in order of decreasing similarity): an unpeeled, overripe kiwi (fuzzy, slightly rough, soft, squishy); an unpeeled, spoiled orange (rounded, slightly soft, textured);}\\[1ex]
Object parts ranked in decreasing hardness: 1 > 2 > 3\\}
}
\end{examplebox}
\caption{Example User Interaction for the Sorting Task.}
\label{fig:SortExample}
\end{figure}

\subsection{Sorting}
The sorting task is designed to demonstrate Octopi-1.5's ability to distinguish tactile properties of objects. This capability has potential use in applications such as sorting/classifying fruits or other items based-on their material properties. 

Given a set of user-chosen items and their corresponding tactile inputs, Octopi-1.5 will be tasked with sorting the items according to properties such as hardness (Fig. \ref{fig:SortExample}). Current experiments show that Octopi is able to sort the balls (100\%) and fruits (93.18\%) but faces difficulty on the unseen objects (43.33\%). We plan to highlight failure cases during the demonstration to show current limitations. This will provide an opportunity to discuss various challenges from annotation (e.g., the relative nature of properties when described using language) to sensing and model training.

\subsection{Free Interaction}
The free interaction segment will allow participants to explore Octopi-1.5's capabilities in a more open-ended manner. Users will be able to:
\begin{itemize}
    \item Provide both visual and tactile inputs to Octopi-1.5.
    \item Engage in an unconstrained chat to ask questions and explore Octopi-1.5's responses.
    \item Test experimental features, including teaching Octopi-1.5 new items.
\end{itemize}
We hope this can help reveal other limitations of the current system and highlight interesting ways users may use Octopi and the TMI.

\section{Conclusion, Limitations, and Future Work}
\label{sec:conclusions}
In summary, we have outlined a series of demonstrations to illustrate the capabilities of Octopi-1.5 and the TMI gripper. We believe the proposed setup is feasible and will be of interest to R:SS attendees, particularly roboticists working on foundation models and tactile robotics. Octopi-1.5 and the TMI showcase interesting features, such as the design of the tactile encoder and the use of RAG, as well as how tactile sensors can be integrated into handheld or robotic grippers. The design and development of such systems remain open questions, and we aim to foster discussions ranging from technical advancements to broader questions about system use and design philosophy.

\paragraph{Limitations.}
Octopi-1.5 is a work-in-progress toward language-conditioned models that enable robots to better utilize vision and touch modalities. While preliminary results are promising, significant limitations remains. General improvements are needed to improve performance on some tasks (e.g., sorting by roughness remains error-prone). This might be addressed by using larger base LLMs, though this requires substantial computational resources. Alternatively, we are exploring more structured decompositions and architectures that leverage prior knowledge about tactile information to mitigate compute/data limitations. We also plan to investigate LTLMs trained to use other sensors such as the TAC-02.

The current RAG module is based solely on text retrieval. Incorporating tactile sample retrieval may further enhance the system, especially for tactile information that is difficult to encode in text. However, more research is needed to isolate what information is best to retrieve in computationally low-cost manner.  

Finally, Octopi-1.5 only outputs text tokens. Directly linking Octopi-1.5 to manipulation tasks is a key area of future work. We are in the process of training a Vision-Tactile-Language-Action (VTLA) model capable of directly outputting robot actions for manipulation tasks, but this is still under development. 

\section*{Acknowledgements}

We gratefully acknowledge Schaeffler Pte. Ltd. for their support and special thanks to Geet Jethwani, Kratika Garg, and Han Boon Siew for their insightful input and assistance throughout the project.
This research is supported by the National Research Foundation, Singapore under its Medium Sized Center for Advanced Robotics Technology Innovation.

\balance 
\bibliographystyle{unsrt}
\bibliography{references}

\end{document}